\documentclass[letterpaper]{article} 
\usepackage{aaai25}  
\usepackage[dvipsnames,table]{xcolor}
\usepackage{times}  
\usepackage{helvet}  
\usepackage{courier}  
\usepackage[hyphens]{url}  
\usepackage{graphicx} 
\usepackage{natbib}  
\usepackage{caption} 
\usepackage{algorithm}
\usepackage{algorithmic}

\usepackage{newfloat}
\usepackage{listings}

\usepackage{booktabs}

\usepackage{multirow}
\usepackage{arabtex}
\usepackage{utf8}
\setcode{utf8}
\usepackage{tcolorbox}

\definecolor{helsinkicolor}{RGB}{230, 230, 250}
\definecolor{florescolor}{RGB}{230, 250, 230}
\definecolor{llamaxcolor}{RGB}{250, 230, 230}
\definecolor{verylightgray}{gray}{0.94} 

\newtcolorbox{helsinkibox}{
    colback=helsinkicolor!30,
    title=Helsinki Translation
}

\newtcolorbox{floresbox}{
    colback=florescolor!30,
    title=Flores-101 Translation
}

\newtcolorbox{llamabox}{
    colback=llamaxcolor!30,
    title=LlamaX Translation
}

\DeclareCaptionStyle{ruled}{labelfont=normalfont,labelsep=colon,strut=off} 
\floatstyle{ruled}
\newfloat{listing}{tb}{lst}{}
\floatname{listing}{Listing}
\title{Bridging Language Barriers in Healthcare: A Study on Arabic LLMs.}
\author{
    Nada Saadi,
    Tathagata Raha,
    Cl{\'e}ment Christophe,\\
    Marco AF Pimentel,
    Ronnie Rajan,
    Praveen K Kanithi
}
\affiliations {M42 Health, Abu Dhabi, UAE}

\begin{document}

\maketitle

\begin{abstract}

This paper investigates the challenges of developing large language models (LLMs) proficient in both multilingual understanding and medical knowledge. We demonstrate that simply translating medical data does not guarantee strong performance on clinical tasks in the target language.  Our experiments reveal that the optimal language mix in training data varies significantly across different medical tasks. We find that larger models with carefully calibrated language ratios achieve superior performance on native-language clinical tasks.  Furthermore, our results suggest that relying solely on fine-tuning may not be the most effective approach for incorporating new language knowledge into LLMs.  Instead, data and computationally intensive pretraining methods may still be necessary to achieve optimal performance in multilingual medical settings.  These findings provide valuable guidance for building effective and inclusive medical AI systems for diverse linguistic communities.

\end{abstract}

\section{Introduction}

The evolution of multilingual language models like Llama3 and GPT-4 marks a significant advancement in natural language processing. However, most LLMs are primarily trained on English and other common European languages, often neglecting low-resource languages with different alphabets, such as Arabic. This limitation poses a significant challenge, particularly in specialized domains like healthcare, where accurate language understanding is crucial.

One major obstacle is the scarcity of high-quality, domain-specific data for these languages. In this paper, we address this challenge by evaluating and improving the capabilities of LLMs for clinical tasks in Arabic. We first conduct a comprehensive evaluation of existing open-source LLMs to assess their performance on medical tasks in both English and Arabic. This analysis provides valuable insights into the current state of LLMs in handling clinical information across different languages.

To further enhance the capabilities of these models, we investigate various techniques, including leveraging existing LLMs for translation, paraphrasing, and generating synthetic data to augment Arabic medical datasets. Specifically, we explore the translation capabilities of both Llama and Qwen, highlighting their strengths and weaknesses in handling medical terminology and nuances in Arabic.

Finally, we fine-tune Llama 3.1 using different mixtures of original and synthetic Arabic medical data. This allows us to analyze the impact of different data sources and augmentation techniques on the model's performance across various clinical tasks. Our findings reveal that the optimal data mixture varies depending on the specific task, emphasizing the importance of careful data curation and augmentation strategies for developing effective clinical LLMs.

Our work focuses on the Llama 3.1 model, but the proposed methodology can be extended to other LLMs, domains, and low-resource languages. We believe this research contributes to developing more inclusive and robust LLMs that can effectively serve diverse linguistic communities and specialized domains.

\section{Related Work}

The emergence of large language models (LLMs) has marked a significant advancement in artificial intelligence, demonstrating impressive capabilities in natural language understanding and generation. Initially developed for general use-cases such as text summarization, translation, and dialogue generation, LLMs have quickly been adopted across diverse industries, including finance, law, and education.

One domain where LLMs have shown considerable promise is healthcare \citep{zhang2023large}.  Recent studies have explored the application of LLMs to a variety of medical tasks, including clinical decision support, medical question answering, and diagnosis assistance.  For instance,  GPT-4 has demonstrated proficiency in medical knowledge evaluation, achieving scores comparable to human experts on standardized medical exams \citep{nori2023capabilities}.  Other models like Meditron \citep{meditron23}, OpenBioLLM \citep{OpenBioLLMs} and Med42 \citep{christophe2024med42} have further advanced the field, with many surpassing GPT-4's performance on specific medical tasks and releasing open-source models usable by the research community and facilitating further advancements in the field.

Evaluating the performance of clinical LLMs, however, presents unique challenges. Most current models are primarily evaluated on question answering tasks using datasets like USMLE, MedQA, and PubMedQA.  New benchmarks such as MEDIC \citep{kanithi2024medic} have emerged to provide a more comprehensive and standardized evaluation of medical LLMs, encompassing tasks like clinical diagnosis, treatment recommendation, and patient education.

Despite the rapid progress, a significant limitation of most existing clinical LLMs is their reliance on English data for training and evaluation. This raises concerns about their applicability and fairness in multilingual healthcare settings, where a significant portion of the population may not be proficient in English.  While multilingual LLMs like Llama, Qwen or Mistral \citep{dubey2024llama, qwen2, jiang2023mistral}  have demonstrated strong performance across multiple languages, adapting these models for both a new language and a specialized domain like healthcare remains an active area of research.  Although some efforts have been made to develop multilingual clinical LLMs \citep{lopez2023gatortronm, wang2024apollo}, broad studies on their cross-lingual performance and generalizability are limited.

Developing LLMs for languages with unique linguistic characteristics and limited digital resources presents additional challenges. Arabic, with its complex morphology, dialectal variations, and relatively scarce medical corpora, exemplifies these challenges. While general-purpose Arabic LLMs like Jais \citep{sengupta2023jais} and Silma \citep{Silma} have emerged, specialized medical models for Arabic remain scarce.  To our knowledge, BiMedix \cite{pieri2024bimedix} is the only model that specifically focuses on building an LLM with multilingual capabilities (Arabic + English) for the healthcare domain.

\section{Evaluating Large Language Models' Capabilities in Arabic Medical Applications}
While many large language models claim to work well in multiple languages and medical tasks, most testing focuses on either general language skills or English medical knowledge separately. Few studies look at how well these models handle medical content in specific non-English languages \citep{10.1145/3589334.3645643}. In this study, we test several popular models of different sizes on Arabic medical benchmarks. Our results show that these models still have a long way to go to match their English performance levels.

\subsection{Arabic Evaluation Datasets}

We utilize a set of Arabic-translated medical datasets for our zero-shot and fine-tuning evaluations. These datasets, originally developed for training and evaluating question-answering systems in the medical domain, include PubMedQA, MedMCQA, MedQA, and Medical MMLU. \textbf{PubMedQA}: This dataset is derived from biomedical research articles \citep{jin2019pubmedqa}. \textbf{MedMCQA} \citep{pmlr-v174-pal22a} and \textbf{MedQA} \citep{jin2021disease} consists of multiple-choice questions coming from Indian and United States medical license exams. Medical \textbf{MMLU} is derived from the MMLU benchmark, specifically focusing on biomedical subsets including Clinical Knowledge, College Biology, College Medicine, Medical Genetics, Professional Medicine, and Anatomy \citep{hendryckstest2021}.

The translation of these datasets into Arabic was conducted using a semi-automated iterative translation pipeline, as detailed in BiMediX \citep{pieri2024bimedix}. This process involves initial translations using language models, followed by human refinement to ensure the accuracy and quality of the translations. The translated datasets maintain the original format of questions and answers, allowing for consistent evaluation across languages.

\subsection{Modifications to Harness Pipeline}

Our research utilizes the Harness evaluation framework \citep{eval-harness}, which calculates log-likelihood scores to evaluate predictive model performance. To accommodate the Arabic language, we made significant modifications to handle its unique attributes, including its distinctive script, complex morphology, and syntactic structure, ensuring accurate processing of Arabic data.

Arabic text runs right-to-left (RTL), unlike English (LTR). We updated the framework to display and process Arabic text correctly. This meant reformatting our dataset while keeping its structure intact.
To conduct zero-shot evaluations in Arabic, we adapted the entire framework for zero-shot testing, converting all prompts, responses, and multiple-choice options to Arabic, ensuring an accurate display and functionality of the Arabic script.
Arabic's linguistic complexity makes context crucial for accurate understanding. A single word can have multiple meanings depending on its grammatical form and context. For instance, the root word \RL{جرب} (j-r-b) can transform into various derivatives that range from "to try" to "to test" to other nuanced meanings, highlighting why machine learning models must carefully consider contextual cues when processing Arabic text.

Additionally, rather than just calculating the probability of generating answer choice labels (e.g., \emph{a}, \emph{b}, \emph{c}, or \emph{d}), we calculate the probability of generating the full answer text. This modification provides a more detailed understanding of the model's performance by taking into account the entire answer generation process.

\begin{table*}
\centering
\begin{tabular}{l|cccccccc}
\toprule
\textbf{Model/Dataset} & \multicolumn{2}{c}{PubMedQA} & \multicolumn{2}{c}{MedMCQA} & \multicolumn{2}{c}{MedQA} & \multicolumn{2}{c}{MMLU} \\
\cmidrule(lr){2-3} \cmidrule(lr){4-5} \cmidrule(lr){6-7} \cmidrule(lr){8-9}
& En & Ar & En & Ar & En & Ar & En & Ar \\
\midrule
Qwen2.5-3B-Instruct \citep{qwen2} & 29.2 & 61.2 & 49.2 & 35.5 & 48.8 & 41.7 & 68.0 & 28.0 \\
Qwen2.5-7B-Instruct \citep{qwen2} & 45.2 & \textbf{74.4} & 56.8 & \textbf{39.5} & 60.2 & 53.9 & \textbf{76.7} & \textbf{34.9} \\
Pangea-7B \citep{yue2024pangeafullyopenmultilingual} & 57.0 & 61.0 & 50.2 & 37.5 & 53.0 & 49.6 & 68.3 & 32.4 \\
Mistral-7B-Instruct\_v0.3 \citep{jiang2023mistral} & 45.8 & 46.6 & 46.3 & 28.0 & 49.3 & 33.8 & 65.1 & 21.6 \\
Llama3.1-8B-Instruct \citep{dubey2024llama} & \textbf{76.2} & 73.2 & \textbf{58.4} & 35.8 & \textbf{62.0} & 29.5 & 73.4 & 46.4 \\ 
Silma-9B-Instruct-v1.0 \citep{Silma} & 75.6 & 64.0 & 54.9 & 38.9 & 61.6 & \textbf{54.7} & 76.1 & 31.5 \\
\midrule
Llama-3.1-70B-Instruct \citep{dubey2024llama} & 73.6 & \textbf{79.4} & 71.8 & 52.2 & 78.2 & 56.6 & \textbf{87.6} & 70.0 \\
Qwen2.5-72B-Instruct \citep{qwen2} & 63.2 & 76.6 & 68.4 & \textbf{56.9} & 76.1 & \textbf{76.1} & 87.4 & 
\textbf{76.1} \\ 
Med42-Llama3.1-70B \citep{christophe2024med42} & 77.6 & 75.0 & \textbf{72.4} & 49.3 & \textbf{80.4} & 53.5 & 86.8 & 67.7 \\
Meditron3-70B \citep{meditron23} & \textbf{80.6 }& 75.8 & 70.9 & 51.2 & 79.3 & 72.0 & 87.0 & 56.6 \\
BiMedix(Bilingual) \citep{pieri2024bimedixbilingualmedicalmixture} & 77.2 & 78.4 & 61.6 & 49.1 & 65.2 & 47.3 & 73.2 & 56.9 \\
\bottomrule
\end{tabular}%
\caption{Accuracy of publicly available models on different Medical QA benchmarks. Even though Llama3.1 models are performing better in English, Qwen2.5 models show a stronger performance in Arabic.}
\label{tab:arab_performance}
\end{table*}

\section{Results}
As shown in Table~\ref{tab:arab_performance}, large language models in all model families exhibit limited performance on Arabic medical benchmarks. While leading models like Llama3.1 achieve high accuracy in English (62.0 and 78.2 on MedQA), their performance significantly degrades when applied to Arabic (29.5 and 56.6). Although Qwen2.5 models demonstrate relatively better performance in Arabic, accuracy remains suboptimal.

We will focus on improving Arabic performance, using Llama3.1 as a case study to explore strategies to achieve English-language proficiency on Arabic medical benchmarks.

\subsection{LLM Adaptation Through Translation Pipeline}

\begin{table*}
\centering
\begin{tabular}{l|cccc}
\toprule
\textbf{Translation Model} & \textbf{PubMedQA} & \textbf{MedMCQA} & \textbf{MedQA} & \textbf{MMLU} \\ 
\midrule
LlamaX \citep{lu2024llamax}& 74.6 & 53.1 & 55.8 & 59.5 \\ 
Helsinki \citep{Helsinki-NLP} & 72.0 & 48.9 & 40.8 & 56.6 \\ 
Flores 101 \citep{flores101} & 72.0 & 36.6 & 31.2 & 34.0 \\ 
Llama3.1-70B-Instruct \citep{dubey2024llama} & 75.8 & 54.8 & 70.5 & 70.7 \\
Qwen2.5-72B-Instruct \citep{qwen2} & \textbf{75.9} & \textbf{55.2} & \textbf{71.3}  & \textbf{71.5} \\
\bottomrule
\end{tabular}
\caption{Performance comparison of various translation models on Arabic medical benchmarks, translated into English and evaluated using Llama3.1-70B-Instruct for accuracy (\%).}
\label{tab:translation_results}
\end{table*}

A straightforward approach to enhance large language model performance across languages is to implement a translation pipeline: convert the Arabic input to English, process it, and translate the output back to Arabic. This method leverages the models' strong English capabilities. However, this approach introduces significant computational overhead, which requires at least three separate model calls instead of one.

We evaluated this pipeline's effectiveness by testing various translation models and comparing Llama-3.1-70B's performance on translated content against its native English capabilities. Using our established evaluation benchmarks, we translated both questions and multiple-choice options before processing them through Llama-3.1-70B, maintaining consistent evaluation methods.

Our investigation included two categories of translation systems: specialized translation models designed for precise Arabic-English conversion, and general-purpose large language models trained on both languages. Though not specifically optimized for translation, these general-purpose models offer broader language understanding.

Our results in Table~\ref{tab:translation_results} reveal that despite their reputation for accuracy, specialized translation models faced considerable challenges with medical content. The nuanced nature of medical terminology makes literal translations problematic, often resulting in technically accurate but contextually inappropriate translations. General-purpose models such as Llama and Qwen demonstrated superior performance in this domain, producing translations that better preserved both technical accuracy and medical context. Although translation pipelines can improve the performance of LLMs in Arabic medical content, the results still lag significantly behind their native English capabilities, highlighting the imperfections of current translation methods. This discrepancy raises serious concerns in the healthcare domain, where accurate understanding and generation of medical information is crucial. Furthermore, the added computational overhead of translation limits the feasibility of deploying such models in resource-constrained environments, hindering their accessibility in regions where they are most needed.

\subsection{Language Specific Finetuning}

We aim to improve large language models' performance by finetuning on bilingual domain-specific data not encountered during pretraining. In this section, we detail our finetuning pipeline, present the datasets utilized, and analyze the optimal balance between English and Arabic training data.

\subsubsection{Finetuning Datasets}

\begin{table*}[htbp]
\centering
\begin{tabular}{lllrr}
\toprule
Dataset & Original & Description & Final & \# of Tokens \\
\midrule
1. AHQAD & Arabic & \begin{tabular}[t]{@{}l@{}}100K sampled based on completeness and\\ paraphrased with Qwen-72B-Instruct\end{tabular} & Arabic & 8.33 M \\
\rowcolor{verylightgray}
2. Translated MED42 Dataset & English & \begin{tabular}[t]{@{}l@{}}500K sampled randomly, cleaned and\\ translated with Qwen-72B-Instruct\end{tabular} & Arabic & 230.69 M \\
3. CIDAR & Arabic & \begin{tabular}[t]{@{}l@{}}10K Instruction-Output good\\ quality dataset\end{tabular} & Arabic & 1.34 M \\
\rowcolor{verylightgray}
4. Med42 Dataset & English & Full English FT dataset & English & 464.97 M \\
5. Synthetic Open-Ended & English & \begin{tabular}[t]{@{}l@{}}\textasciitilde{}200K sampled based on 1-5 rating and\\ translated with Qwen-72B-Instruct\end{tabular} & Arabic & 240 M \\
\midrule
\multicolumn{4}{l}{\textbf{Total \# of Arabic Tokens}} & \textbf{480.36 M} \\
\multicolumn{4}{l}{\textbf{Total \# of English Tokens}} & \textbf{469.97 M} \\
\bottomrule
\end{tabular}%
\caption{Dataset Overview with Language Distribution}
\label{tab:dataset-overview}
\end{table*}

Due to the scarcity of high-quality Arabic clinical data, we developed a comprehensive data preparation pipeline. This section details our methodology for cleaning existing datasets, generating new data, and performing translations to ensure robust data quality. The size of each dataset is described in Table~\ref{tab:dataset-overview}.

\begin{itemize}
    \item \textbf{Arabic Health Questions \& Answers Dataset (AHQAD)}: The AHQAD dataset, with its 90 richly diverse categories, offers a comprehensive landscape of medical and healthcare-related themes tailored specifically for the Arabic-speaking region. This collection spans an extensive array of topics, from general medicine to specialized fields such as cardiology, pediatrics, and oncology, as well as practical areas like pharmacology and patient care. It also includes emergent fields such as telemedicine and health informatics. 

    For the AHQAD dataset, which comprises medical queries, we applied a more selective filtering process. Out of its total 298,000 entries, we chose 100,000 that featured the most complete questions. This was crucial as it ensured the data's clarity and relevance, enhancing the quality of the training material. Responses were standardized using Qwen2.5-72B model for paraphrasing. We instructed the model to rewrite responses while maintaining the original meaning, removing typos and complex abbreviations, and improving clarity. The model was explicitly prompted to preserve the original information without adding any new content.

    \item \textbf{Translated Med42 Dataset}: We leverage the finetuning dataset used for finetuning Med42-v2. The Med42 dataset is curated from various medical and biomedical resources and it also features chat interactions and chain-of-thought reasoning apart beyond simple question-answering. 

    For the Med42 dataset, we randomly selected 500,000 records from the dataset used to train Med42 \cite{christophe2024med42}. This dataset was then translated to Arabic using the Qwen2.5-72B-Instruct model, chosen for its strong performance on Arabic benchmarks, as demonstrated in Table~\ref{tab:arab_performance}. Our manual evaluation further confirmed that its translations are of higher quality, with superior context preservation, both in medical and non-medical scenarios. Following translation, we meticulously cleaned the data to ensure only Arabic samples were retained, discarding any residual English phrases, ultimately preserving approximately 90\% of the dataset. 
    
    \item \textbf{Culturally Relevant Instruction Dataset For Arabic (CIDAR)}: The CIDAR dataset \cite{alyafeai2024cidarculturallyrelevantinstruction} on Hugging Face is an instruction-output pair dataset explicitly crafted for Arabic NLP tasks, making it particularly valuable for training models in zero-shot and few-shot learning scenarios. Each record in the dataset features a distinct instruction—a prompt, question, or directive in Arabic—and a corresponding output that provides a precise, contextually relevant response. This structure is intended to help models interpret and generate responses across various types of tasks, such as factual questions, conversational exchanges, and directive-based commands, thereby enhancing the model's instruction-following capabilities.

    \item \textbf{Synthetic Open-Ended QA}: The preparation of healthcare-specific synthetic question-answer pairs employs a systematic multi-stage approach. The process begins by randomly selecting seed questions from HealthSearchQA, ExpertQA, and MedicationQA datasets. These seed questions serve as the foundational examples for iteratively building a synthetic instruction dataset. The iterative process involves using the seed instructions as few-shot examples to generate new synthetic instructions. With each iteration, the pool of instructions expands, ensuring diversity and coverage across healthcare topics. Importantly, to preserve data independence, the final synthetic instruction dataset excludes the original seed instructions. The generated questions are then processed using Llama-3.1-70B-Instruct to create comprehensive responses. To ensure quality, these responses undergo evaluation using the same model on a scale of 1 to 5, with only pairs rated 5 being retained in the final dataset to ensure high quality. The entire dataset is subsequently translated into Arabic using Qwen2.5-72B-Instruct.
\end{itemize}

\begin{table*}[h!]
\centering
\begin{tabular}{lccccccccc}
\toprule
\textbf{Model Configuration} & \textbf{\begin{tabular}[c]{@{}c@{}}Dataset Ratio\\ (Arabic-English)\end{tabular}} & \multicolumn{8}{c}{\textbf{Accuracy}} \\

\cmidrule(l){3-10}
& & \multicolumn{2}{c}{PubMedQA} & \multicolumn{2}{c}{MedMCQA} & \multicolumn{2}{c}{MedQA} & \multicolumn{2}{c}{MMLU} \\
\cmidrule(lr){3-4} \cmidrule(lr){5-6} \cmidrule(lr){7-8} \cmidrule(lr){9-10}
& & En & Ar & En & Ar & En & Ar & En & Ar \\
\midrule
\rowcolor{gray!10}
\textbf{Llama 3.1 8B} & Baseline & 38.0 & 34.4 & 50.0 & 32.2 & 55.1 & 27.3 & 64.8 & 39.8 \\
1. Arabic Only & 100\%-0\% & 68.6 & \textbf{71.2} & 56.6 & 32.2 & 55.8 & 27.5 & 72.5 & 40.4 \\
2. Strong Arabic Majority & 80\%-20\% & \textbf{73.6} & 68.0 & 56.7 & \textbf{35.1} & 57.4 & 27.5 & 71.1 & 40.2 \\
3. Arabic Majority & 60\%-40\% & 69.2 & 63.0 & 56.6 & 32.9 & 57.9 & 28.3 & \textbf{72.9} & 40.9 \\
4. Balanced Distribution & 50\%-50\% & 70.0 & 61.2 & 56.6 & 34.7 & \textbf{58.7} & 29.5 & 71.9 & 42.3 \\
5. English Majority & 40\%-60\% & 59.0 & 53.8 & 57.0 & 33.4 & 57.7 & \textbf{29.8} & 70.8 & 42.0 \\
6. Strong English Majority & 20\%-80\% & 67.0 & 53.8 & 57.6 & 33.4 & 57.3 & 28.7 & 71.7 & 42.0 \\
7. English Only & 0\%-100\% & 72.8 & 61.2 & \textbf{58.8} & 34.7 & 58.5 & 29.5 & 71.62 & \textbf{42.4} \\
\midrule
\rowcolor{gray!10}
\textbf{Llama 3.1 8B-Instruct} & Baseline & 76.2 & \textbf{73.2} & 58.4 & 35.8 & \textbf{62.0} & 29.5 & 73.4 & \textbf{46.4} \\
1. Arabic Only & 100\%-0\% & 72.0 & 72.0 & 58.7 & 31.9 & 57.9 & 29.2 & 73.6 & 30.0 \\
2. Strong Arabic Majority & 80\%-20\% & 76.6 & 69.0 & 58.4 & 34.7 & 59.6 & 29.7 & 72.8 & 42.1 \\
3. Arabic Majority & 60\%-40\% & \textbf{76.8} & 68.4 & 58.1 & 33.7 & 60.3 & 30.9 & 73.1 & 43.0 \\
4. Balanced Distribution & 50\%-50\% & 71.2 & 66.2 & 58.7 & 34.6 & 60.2 & \textbf{33.5} & \textbf{73.7} & 42.5 \\
5. English Majority & 40\%-60\% & 74.8 & 66.2 & 59.1 & \textbf{36.8} & 61.9 & 33.5 & 72.3 & 42.1 \\
6. Strong English Majority & 20\%-80\% & 73.4 & 64.0 & \textbf{59.5} & 35.1 & 60.8 & 31.6 & 73.3 & 42.8 \\
7. English Only & 0\%-100\% & 68.4 & 60.0 & \textbf{59.5} & 36.4 & 60.6 & 30.2 & 73.6 & 46.3 \\
\midrule
\rowcolor{gray!10}
\textbf{Llama 3.1 70B} & Baseline & 15.6 & 51.8 & 65.1 & 41.6 & 75.9 & 48.2 & 82.4 & 55.8 \\
1. Arabic Only & 100\%-0\% & 70.6 & \textbf{76.8} & 68.8 & \textbf{51.7} & 75.3 & \textbf{53.3} & 84.3 & \textbf{67.3} \\
2. Balanced Distribution & 50\%-50\% & \textbf{78.0} & 55.0 & 68.9 & 50.5 & \textbf{76.9} & 50.9 & 84.0 & 59.3 \\
3. English Only & 0\%-100\% & 75.2 & 48.6 & \textbf{70.2} & 47.5 & 76.1 & 48.6 & \textbf{85.7} & 61.9 \\
\midrule
\rowcolor{gray!10}
\textbf{Llama 3.1 70B-Instruct} & Baseline & 73.6 & \textbf{79.4} & \textbf{71.8} & 52.2 & \textbf{78.2} & \textbf{56.6} & \textbf{87.6} & \textbf{70.0} \\ 
1. Arabic Only & 100\%-0\% & \textbf{79.8} & 78.2 & 70.6 & \textbf{52.8} & 77.1 & 55.8& 86.3 & 67.1 \\
2. Balanced Distribution & 50\%-50\% & 77.2 & 61.8 & 71.4 & 50.8 & 76.2 & 55.6 & 86.9 & 67.0 \\
3. English Only & 0\%-100\% & 75.0 & 48.2  & 71.7 & 49.7 & 77.8 & 52.9 & 87.5 & 66.5  \\
\bottomrule
\multicolumn{10}{l}{\small \textit{Note: Results show performance on English (En) and Arabic (Ar) evaluations for each metric.}} \\
\end{tabular}
\caption{Accuracy of Finetuned Llama3.1-8b and 70b models with Different Arabic-English Dataset Ratios on medical QA benchmarks. Fine-tuning Llama 3.1 models on varying Arabic-English dataset ratios yields inconsistent results across medical QA tasks.  Even large instruct models show limited improvement on Arabic benchmarks after fine-tuning.}
\label{tab:combined-results}
\end{table*}

\subsubsection{Fine-tuning Pipeline}
The bilingual medical fine-tuning pipeline explores the optimal combination of Arabic and English medical datasets to enhance model performance on both English and Arabic medical tasks. The pipeline incorporates two distinct data streams: high-quality Arabic medical content obtained through rigorous cleaning and filtering of native Arabic medical datasets, and carefully translated English medical datasets that maintain clinical accuracy in both languages. For different ratios, we maintain a constant number of 469.97M tokens, randomly sampling from our dataset presented in Table~\ref{tab:dataset-overview}.

We finetuned both Llama3.1 models. We employ the classic auto-regressive loss for finetuning. Loss is backpropagated only on output tokens. This approach ensures that the model learns to generate appropriate responses and not learn to generate the prompts. Our training samples are concatenated into chunks of 8192 tokens. Each model was finetuned for two epochs over our curated dataset using a cosine learning rate schedule between $1\times10^{-5}$ and $1\times10^{-6}$. All experiments are performed on a cluster of 4 H100 nodes.

\subsubsection{Results}

Our results in Table~\ref{tab:combined-results} show that different ratio of Arabic-English data yield to different performance levels depending on the evaluation task. For PubMedQA, training with exclusively Arabic data produces the best accuracy (71.2). While for MedMCQA and MedQA, the models perform best with strong Arabic majority and English Majority, respectively (35.1 and 29.8). Surprisingly, for the MMLU datasets, which focuses on testing direct knowledge application, using only English data, achieves 42.4 compared to the 39.8 zero-shot accuracy.

These patterns remain consistent across both the base and instruct models. These results highlight the fact that the relationship between the language distribution used for finetuning and performance is fundamentally linked to the nature of each task. For instance, PubMedQA requires complex analytical reasoning within medical contexts, while MMLU focuses on structured knowledge assessment through multiple-choice questions. This difference suggests that tasks requiring a deeper understanding of context need stronger language-specific training.

Interestingly, our findings on the larger 70B parameter models show more consistent behavior across all Arabic tasks, with Arabic-only training data consistently achieving the best results. This suggests that larger models may handle language-specific tasks more uniformly than their smaller counterparts, given their greater capacity to abstract and generalize linguistic features across different training distributions. Llama3.1-70B base model exhibits poor performance on the PubMedQA test set due to its inability to follow the chat-template for a highly context-based task. Intriguingly, our fine-tuned version of Llama3.1-70B-Instruct rarely outperforms the original model, suggesting that it has reached its maximum capabilities after subsequent pretraining, supervised fine-tuning, and alignment stages.

\section{Conclusion}

Our research into Arabic-English medical AI reveals critical insights for developing truly effective multilingual language models.

First, we highlight the significant performance gap between English and Arabic, especially pronounced in smaller models. This disparity underscores the need for models deeply trained in specific languages to achieve genuine language understanding and complex medical reasoning.   Smaller models, with their limited capacity, struggle to capture the nuances of different languages and medical terminology, resulting in a substantial performance gap between languages like English and Arabic.

Second, while some general language models demonstrate superior translation capabilities compared to specialized translation models, they come with high computational costs and are not perfect. General language models, despite their broader training data, still have limitations in accurately translating medical terminology and complex linguistic structures, highlighting the need for further research in this area.

Third, fine-tuning models do not always guarantee improved performance compared to the baseline, and the results are highly dependent on the distribution of languages in the training data. The effectiveness of fine-tuning can vary significantly depending on the specific language mix used in the training data, suggesting that a careful balance of languages is crucial for optimal performance.

We acknowledge that our evaluation primarily focuses on close-ended question benchmarks, which, while valuable for assessing domain knowledge, do not fully capture the generation capabilities, safety, and bias aspects of a model. These aspects are crucially important for any healthcare model. Therefore, we advocate for new benchmarks, such as MEDIC \citep{kanithi2024medic}, to include multilingual capabilities tests to address these critical dimensions.

It is important to note that the impact of language mixing is particularly significant when dealing with languages that have vastly different alphabets, such as Arabic, Chinese, or Latin-based languages. The non-overlapping nature of their tokens can lead to unique challenges in training and optimization. Moreover, the performance of a model in one domain should ideally transfer seamlessly across multiple languages. It should be easier for a model to learn technical vocabularies in a new language if it is already trained on that domain and possesses a good understanding of the language. Therefore, the transfer capabilities of a model for a specific domain from one language to another should be high.

Thus, we need to continue relying on extensive pre-training for models to learn a new language effectively. At the same time, exploring the transfer capabilities of models for specific domains across languages is crucial. The ultimate goal extends beyond technical achievement: we aim to create AI systems that can break down language barriers, provide accurate medical insights, and expand healthcare access, especially in underserved and linguistically diverse communities. This means developing models that do not just translate words, but truly comprehend the intricate cultural and linguistic subtleties of medical communication. Achieving this goal will require models that can not only translate medical information accurately but also understand the cultural context and linguistic nuances associated with different languages, ensuring effective communication and healthcare access for diverse populations.

\bibliography{aaai25}

\appendix
\section{Appendix}
\subsection{Translation Comparison for Medical Terminology}

\begin{helsinkibox}
\textbf{Original Translation:}
\begin{quote}
How big is the bottle you're gonna use in a patient who needs a quick blood transfusion (based on medical knowledge in 2020)?
\end{quote}

\textbf{Options:}
\begin{itemize}
    \item 18 Gg.
    \item 20 Gs.
    \item 22 Gig.
    \item 24 Gg.
\end{itemize}
\end{helsinkibox}

\begin{floresbox}
\textbf{Original Translation:}
\begin{quote}
What size of cannula would you use in a patient who needed a rapid blood transfusion (as of 2020 medical knowledge)?
\end{quote}

\textbf{Options:}
\begin{itemize}
    \item 18 gauge
    \item 20 gauge
    \item 22 gauge
    \item 24 gauge
\end{itemize}
\end{floresbox}

\begin{llamabox}
\textbf{Original Translation:}
\begin{quote}
What is the volume of the cannula you will use for a patient who needs a rapid blood transfusion (according to medical knowledge in 2020)?
\end{quote}

\textbf{Options:}
\begin{itemize}
    \item 18 qij.
    \item 20 qij.
    \item 22 qij.
    \item 24 qij.
\end{itemize}
\end{llamabox}

\section{Analysis}

\begin{tcolorbox}[title=Translation Quality Comparison]
\begin{itemize}
    \item \textbf{Helsinki:} Uses informal language ("gonna") and confuses the medical terminology by referring to a "bottle" instead of a cannula. Unit notation is inconsistent (Gg., Gs., Gig.).
    
    \item \textbf{Flores-101:} Provides the most accurate medical terminology, using "cannula" and "gauge" correctly. Maintains consistent formatting and professional medical language.
    
    \item \textbf{LlamaX:} Uses correct medical term "cannula" but focuses on "volume" rather than size. Unit notation shows consistent but incorrect translation (qij.).
\end{itemize}
\end{tcolorbox}

\begin{tcolorbox}[colback=blue!5,colframe=blue!40!black,title=Qwen Translation Example]
\textbf{Input Context:}

Original English Text:
\begin{quote}
Why do some people develop hypothyroidism after radioactive iodine treatment for thyroid cancer? Radioactive iodine treatment for thyroid cancer can sometimes lead to decreased thyroid function in some individuals. This occurs due to the damage caused by radiation to the thyroid gland cells.
\end{quote}

\textbf{Qwen Translation:}

\begin{arabtext}
لماذا يعاني بعض الأشخاص من نقص نشاط الغدة الدرقية بعد العلاج الإشعاعي لسرطان الغدة الدرقية؟العلاج الإشعاعي لسرطان الغدة الدرقية قد يؤدي في بعض الأحيان إلى فرط نقص الغدة الدرقية في بعض الأفراد. يحدث هذا بسبب الضرر الذي يسببه الإشعاع للخلايا في الغدة الدرقية.
\end{arabtext}
 
\end{tcolorbox}

\end{document}